\documentclass[11pt,twoside]{article} 
\usepackage{acta-info}
\usepackage{euler}

\usepackage[T1]{fontenc}
\usepackage[english]{babel}
\usepackage{float}
\usepackage{textcomp}
\usepackage{amsmath}
\usepackage{stackrel}
\usepackage{graphicx}
\usepackage{gensymb}

\newcommand{\vol}{\textbf{x,} y (201z) xx--yy}   
\newcommand{\amss}{\amssubj{%
68U99
}}     
\newcommand{\CRs}{\CRsubj{%
I.4.5
}}   
\newcommand{\keyw}{\keywords{%
computer vision, triangulation, reconstruction
}}

\begin{document}

\title{%
Comparing two- and three-view Computer Vision
}

\maketitle

\oneauthor{%
\href{http://www.domain.edu}{Zsolt Levente KUCSV\'AN} 
}{%
\href{http://www.domain.edu}{Sapientia Hungarian University of Transylvania\\ T\^argu Mure\c s, Romania}
}{%
 \href{mailto:kzsolt@student.ms.sapientia.ro}{kzsolt@student.ms.sapientia.ro}
}




\short{%
Zs. Kucsv\'an
}{%
Comparing two- and three-view Computer Vision
}

\begin{abstract}
To reconstruct the points in  three dimensional space, we need at least two images. In this paper we compared two different methods: the first uses only two images, the second one uses three. During the research we measured how camera resolution, camera angles and camera distances influence the number of reconstructed points and the dispersion of them. The paper presents that using the two-view method, we can reconstruct significantly more points than using the other one, but the dispersion of points is smaller if we use the three-view method. Taking into consideration the different camera settings, we can say that both the two- and three-view method behaves the same, and the best parameters are also the same for both methods.
\end{abstract}


\section{Introduction}

Computer Vision plays an increasingly important role in our days. It's use is very diverse. The diversity is reflected in it's use from engineering to medical applications but Computer Vision also plays a major role in the entertainment industry. Some examples of it's usage are: autonomous cars, face detection, three dimensional triangulation, extended reality, Google Street View etc.
\\
Most of the applications mentioned above require a variety of tools. Autonomous cars use different capture devices, such as LIDAR (laser-based sensor), see for instance \cite{huang2009novel}, to achieve \textquotedbl vision\textquotedbl . Likewise, there is a need for a device (eg. MRI or CT) that can capture an image from inside the body to segment the tumors. It is noticeable that similarly to the complex structure of the human visual system, for the computer vision, we also need a complex system of physical devices. Of course, this complex physical system is not enough. There is also a need for effective software, that processes and interprets the information gained through the devices.
\\
From a large set of applications, the three-dimensional triangulation
may seem to be the simplest. For this application, we do not need
anything else but just different photos about the same object. It
is not necessary for images to be made with the same camera and it
is not important how the cameras are placed when capturing the images,
but it is important to have at least one common object on the images.
Based on these, it seems this problem can be solved more easily than
the rest. Nevertheless, if we are getting deeper in this subject,
it turns out that although this task is associated with the least
constraint, it is one of the most difficult to solve mathematically
and to write optimal software for this.
\\
In this paper the main goals are to compare the two-view triangulation with the three-view one and to create a data set, that is suitable for the previously mentioned purpose. Comparing the two methods, we had the following aims: to compare
the two methods based on input images with different resolution, in
the case of different angles of the cameras and finally in the case
of different distances of the cameras.
\\
In order to make the right measurements, the following are required:
to identify common pixels on images, to calculate the camera matrices
based on the images and to triangulate the identified common pixels
based on the camera matrices (using both two-view and three-view method).

\section{Mathematical background and previous results}

1997 is a significant year in the history of Computer Vision. At this
point, there were methods for the triangulation problem, but in this
year Richard I. Hartley with Peter Sturm published a method, which
gives an optimal solution in case we are using image pairs. Their
method differs from previous methods, so not the algebraic error was
minimized but the geometrical. By this method they achieved to solve
the problem optimally by finding the roots of a 6th grade polynomial, see \cite{hartley1997triangulation}.
\\
In 2005 Henrik Stewenius, Frederik Schaffalitzky
and David Nister published a paper about solving optimal the three-view
triangulation. This method is based also on minimizing the geometric
error, but instead of the 6th grade polynomial we have to find the
roots of a 47th grade polynomial to get the optimal solution, see \cite{stewenius2005hard}.
\\
The three-view triangulation compared to the two-view has just one
more input (the third picture) but the problem to find the optimal
triangulation has become more complicated. This is also a question,
that if we want to triangulate optimally from $n$ pictures, how complicated
will be the polynomial we have to solve. We don't know for sure the
answer to this question, but there is a paper from 2016 in which there
is formulated a conjecture about this. Conform to the conjecture,
if we have $n$ pictures, the grade of polynomial will be the
following (see \cite{draisma2016euclidean}):

\[
\frac{9}{2}n^{3}-\frac{21}{2}n^{2}+8n-4
\]

\subsection{Camera coordinate system}

The corresponding point of $X_{3D}=(x_{1},x_{2},x_{3})^{T}$ is the
$X_{2D}=(x_{1},x_{2})^{T}$ in the camera coordinate system. This
has the following form in homogeneous coordinates (see \cite{computervision}):

\begin{equation}
\begin{pmatrix}x_{1}\\
x_{2}\\
x_{3}\\
1
\end{pmatrix}\longmapsto\begin{pmatrix}fx_{1}\\
fx_{2}\\
x_{3}
\end{pmatrix}=\begin{pmatrix}f & 0 & 0 & 0\\
0 & f & 0 & 0\\
0 & 0 & 1 & 0
\end{pmatrix}\begin{pmatrix}x_{1}\\
x_{2}\\
x_{3}\\
1
\end{pmatrix}\label{eq:1}
\end{equation}
\\
Transformation \eqref{eq:1} is a simple product
of matrices. The left-hand side of the product is the projection matrix,
that contains the parameters of the camera. These parameters determine
the formation of the image. The right-hand side of the product is
the three-dimensional point itself, which we would like to take picture
of. The actual projection matrix looks like the following (see \cite{computervision}):

\begin{equation}
P=\begin{pmatrix}s_{x}f & a & s_{x}o_{x} & 0\\
0 & s_{y}f & s_{y}o_{y} & 0\\
0 & 0 & 1 & 0
\end{pmatrix},\label{eq:2}
\end{equation}
where $f$ is the focal length, $(s_{x},s_{y})$ the zoom of the camera,
$(o_{x},o_{y})$ the translation of the axes of the camera coordinate
system and $a$ determines the shape of the pixels. The parameter
$a$ is not null just in the case, where the axes of the coordinate
system are not perpendicular to each other, see \cite{computervision}.
\\
The projection matrix has also the following form:

\begin{equation}
P=K[I|O],\label{eq:3}
\end{equation}
where $K$ is the calibration matrix, see \cite{computervision}.

\subsection{World coordinate system}

It is not enough to consider just the camera coordinate system, because
the real points which we want to take pictures of are in the world
coordinate system. The connection of the two coordinate systems are
the following:
\begin{enumerate}
	\item There is a translation $t$ between them
	\item The world coordinate system is rotated 
\end{enumerate}
Taking into consideration these, the projection matrix has the following
form:
\begin{equation}
P=K[R|Rt]=KR[I|t]=K[R|-R\tilde{C}]=KR[I|-\tilde{C}],
\end{equation}
where $R$ is the rotation and, $t$ is the translation. $\tilde{C}$
is inhomogeneous coordinate of camera center in the world coordinate
system ($t=-\tilde{C}$), see \cite{computervision}.

\subsection{Camera calibration}

From \eqref{eq:1}, we can easily determine the image of a
point if projection matrix $P$ is known. The process of determining
the projection matrix $P$ is called camera calibration. 
\\
One method to this process is using calibration pattern. The corners
of a chessboard can be easily identified, so they are often used. 
The most used calibration pattern is the Tsai grid, see \cite{tsai1986efficient}.
\\
$P$ has 12 elements, but it's degree of freedom is just 11, so we
need 11 equations to determine the projection matrix. 
\\
Because of the noise of the images, the 11 equations won't give an
exact solution, so we use more equations and minimize the algebraic
error of the over-defined equation system, see \cite{computervision}.
\begin{equation}
Ap=0,\label{eq:4}
\end{equation}
where $A$ has size $2N\times12$ and $p=(P_{11},P_{12},\,...\,,P_{34})^{T}$
contains the unknown elements of the projection matrix ($P_{ij}\in P,\:i=\overline{1,3},\:j=\overline{1,4}$). 

\subsection{Geometrical error}

\eqref{eq:4} minimizes the algebraic error of the system,
but this won't be optimal geometrically. For minimizing the geometrical
error of $n$-view method, we use the following equation, from \cite{computervision}:
\begin{equation}
Err = \sum_{i=1}^{n}\left\|X_{2D_{i}}-PX_{3D_{i}}\right\|^{2},
\end{equation}	
\\
where $X_{2D}$ is the coordinate of a pixel on the image, and $PX_{3D}$ is coordinate projected by the camera matrix. This is a least squares problem, which we can solve for example with \textit{Levenberg-Marquard algorithm}, see \cite{hartley2003multiple}.
\section{Practical implementation}
To make the proper measurements, we needed an appropriate set of data. Since we did not find a suitable data set, we generated one, see \cite{imgs}. The pictures for this data set were made with a Google Pixel 2 smartphone, using the OpenCamera (see \cite{openCamera}) application freely available for Android phones. The specifications of the camera are the following:
\begin{center}
	\begin{table}[H]
		\begin{centering}
			\begin{tabular}{|c|c|}
				\hline 
				Sensor type & CMOS\tabularnewline
				\hline 
				Sensor size & 1/2.6\textquotedbl\tabularnewline
				\hline 
				Aperture & f/1.8\tabularnewline
				\hline 
				Focal length & \ensuremath{\approx}4.47 mm\tabularnewline
				\hline 
				Image Resolution & 4032x3024 (12.19 MP)\tabularnewline
				\hline 
				Pixel size & 1.4 \textmu m\tabularnewline
				\hline 
			\end{tabular}
			\par\end{centering}
		\caption{Google Pixel 2 - camera specifications\cite{pixel2}}
	\end{table}
	\par
\end{center}

\begin{figure}[h!]
	\begin{centering}
		\includegraphics[angle=90,scale=0.50]{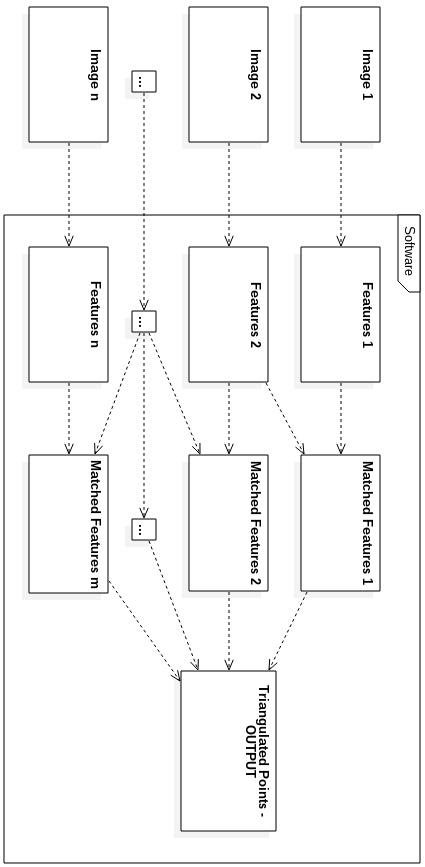}
		\par\end{centering}
	\caption{OpenMVG Reconstruction System}
\end{figure}

\subsection{OpenMVG library}

For comparing the two methods, we used the OpenMVG open-source library, see \cite{openMVG}. This library is designed for computer-vision scientists. OpenMVG provides solutions for multiple problems in Computer Vision, such as the triangulation. 

\section{Evaluation and conclusion}

Given the number of found common pixels, the two-view method founds $1.45-1.86$
times more common pixels than three-view one, but there are some questions. The
first question would be, what happens with common pixels on all images?
In this case the two-view method does three triangulations: using
images $1-2$, $1-3$ and $2-3$. From these three triangulations two are redundant.
Furthermore if there is noise on the images, the needlessly triangulated
points won't coincide with each other, however they should.

\subsection{Evaluation of different resolution measurements}

\begin{figure}[h!]
	\centering
	\includegraphics[scale=0.241]{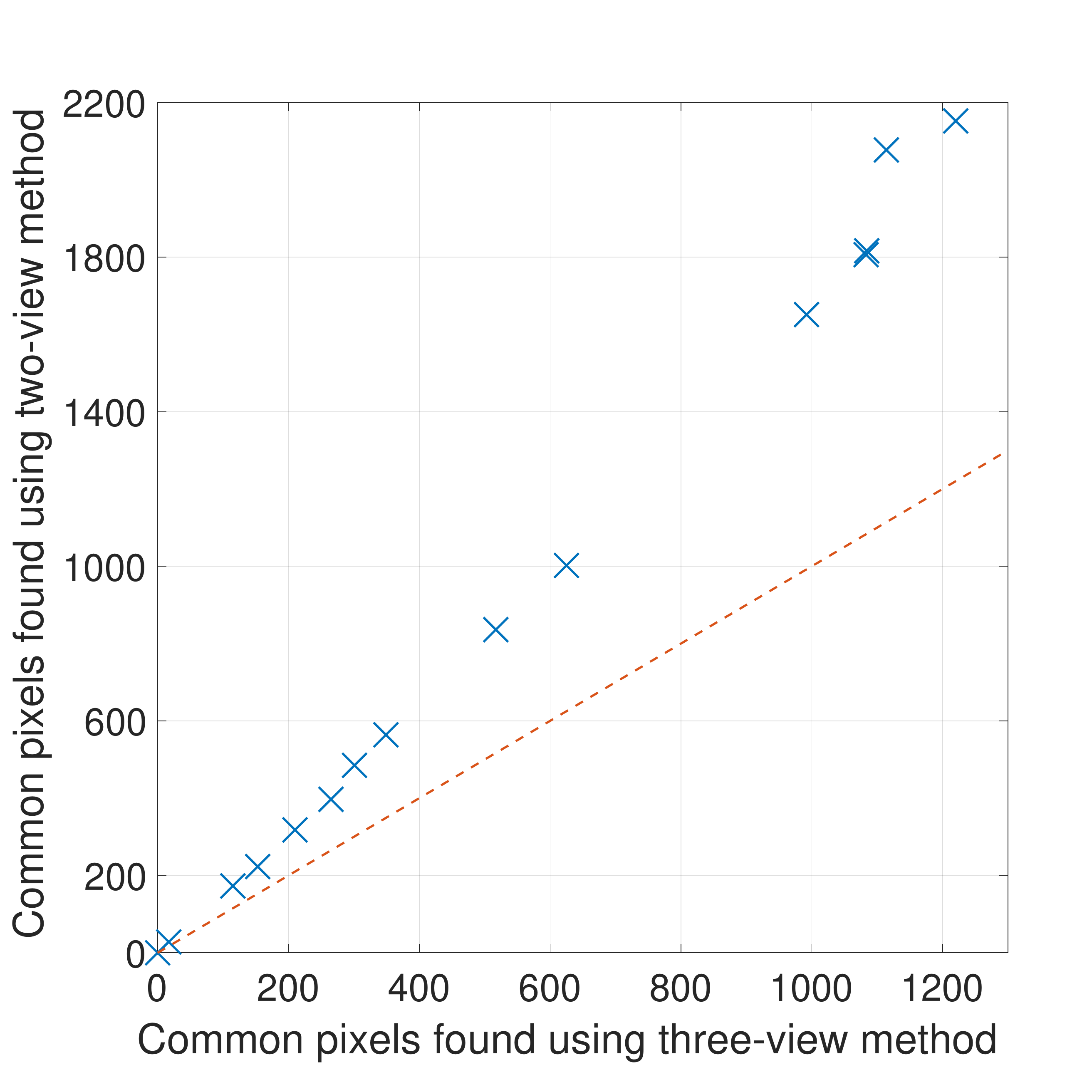}
	\includegraphics[scale=0.241]{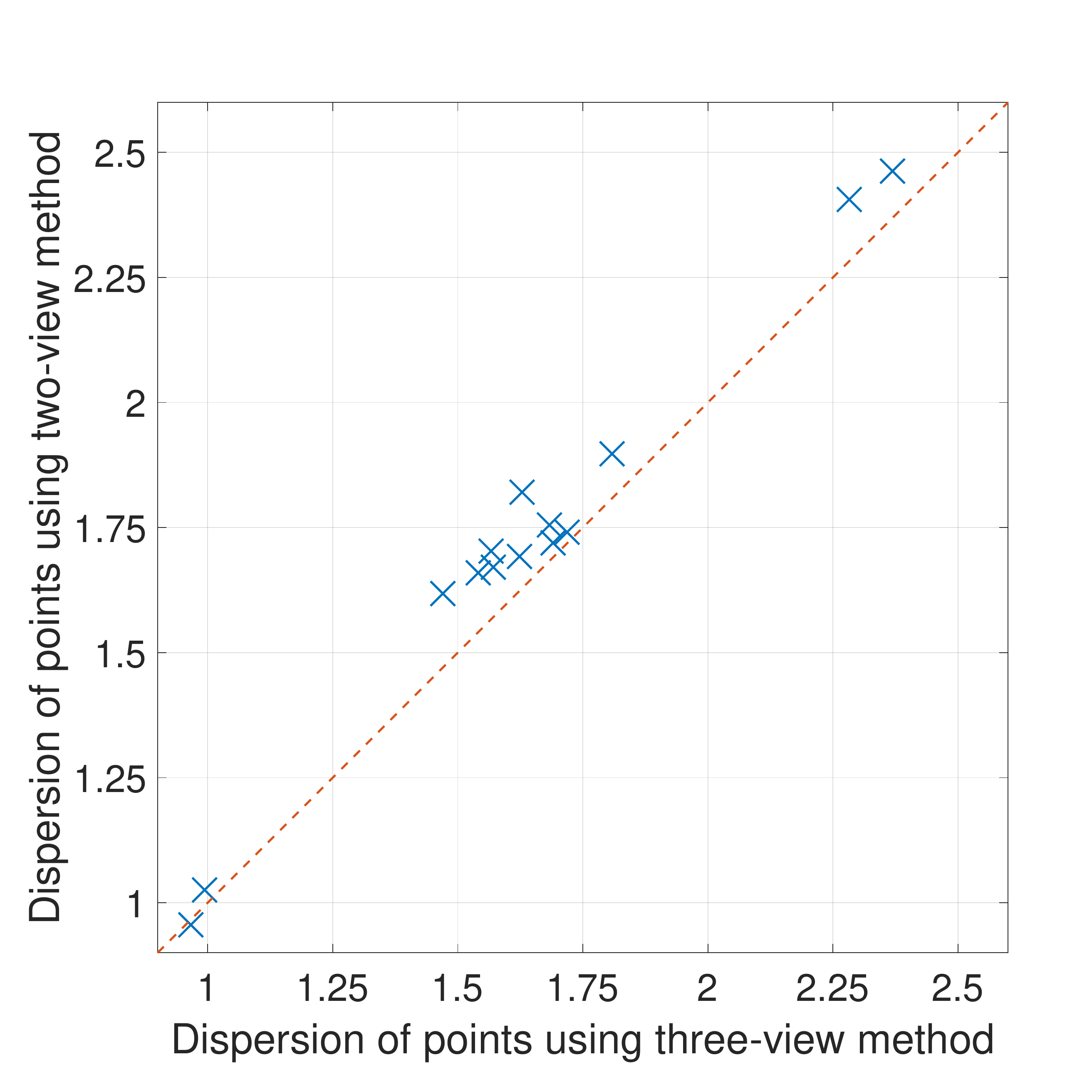}
	\caption{Different resolution measurements}
	\label{fig:resolutions}
\end{figure}

We can notice, that the smaller the resolution, the less common points
will find the software. At very low resolution $(480\times320)$, the software
does not work at all. Up to FullHD resolution $(1920\times1080)$ the software
finds less than $1,000$ while above FullHD resolution finds thousands
of common pixels with the two-view method. The three-view method has
similar results. Up to UltraHD resolution $(3840\times2160)$ the number of
reconstructed pixels increases steadily, while at the highest resolution
$(4032\times3024)$ a larger drop is detected ($426$ points for two-view and
$122$ for three-view method). One possible explanation for this is the
higher noise entering at high resolution. If there was no noise on
the images, the number of reconstructed points would increase steadily
by increasing the resolution.\\
Analyzing the dispersion, we can observe for the three-view method
this is smaller. The dispersion at three-view method can be smaller
from $1.2\%$ up to $11.8\%$. On average, the dispersion of three-view
triangulation is less by $5.6\%$ than the two-view method. \\

\subsection{Evaluation of different angles measurements}

\begin{figure}[h!]
	\centering
	\includegraphics[scale=0.3]{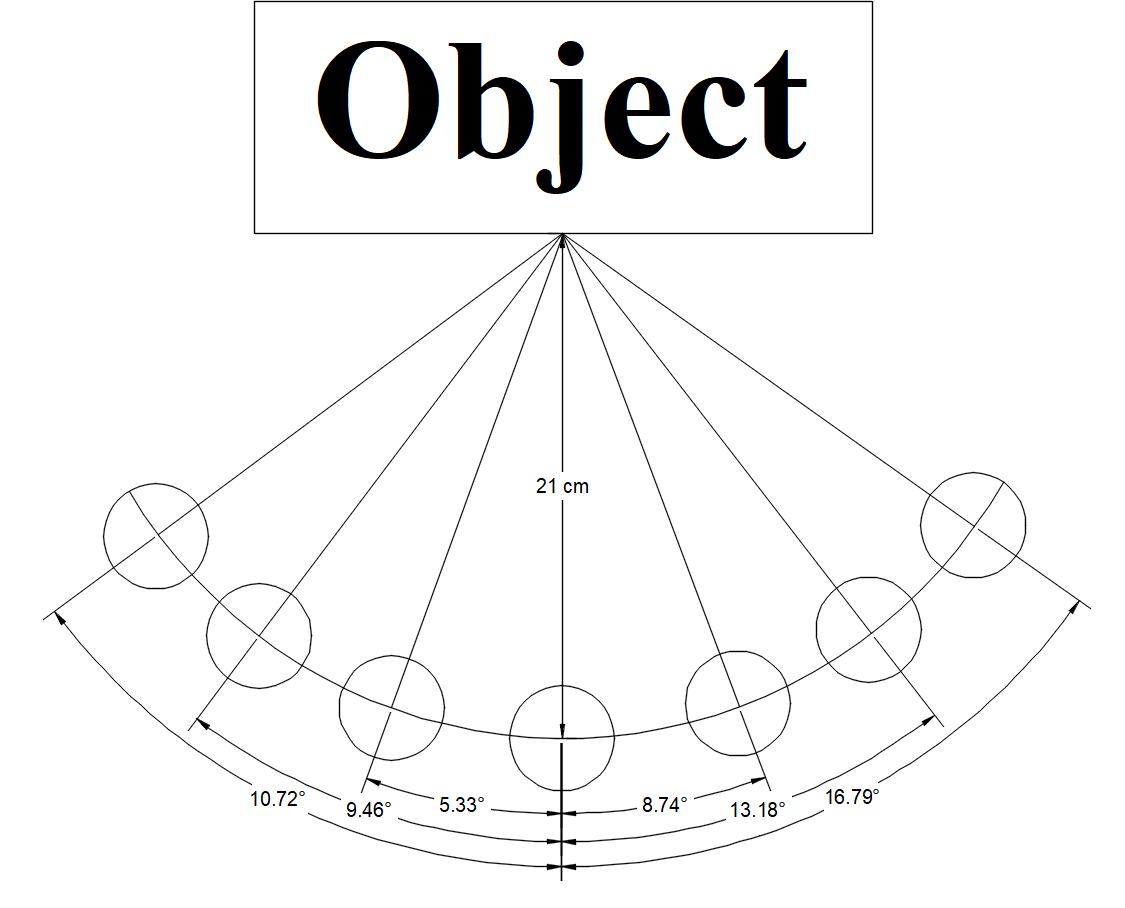}
	\caption{Camera positions for different angle measurements}
	\label{fig:ang}
\end{figure}

\begin{figure}[h!]
	\centering
	\includegraphics[scale=0.241]{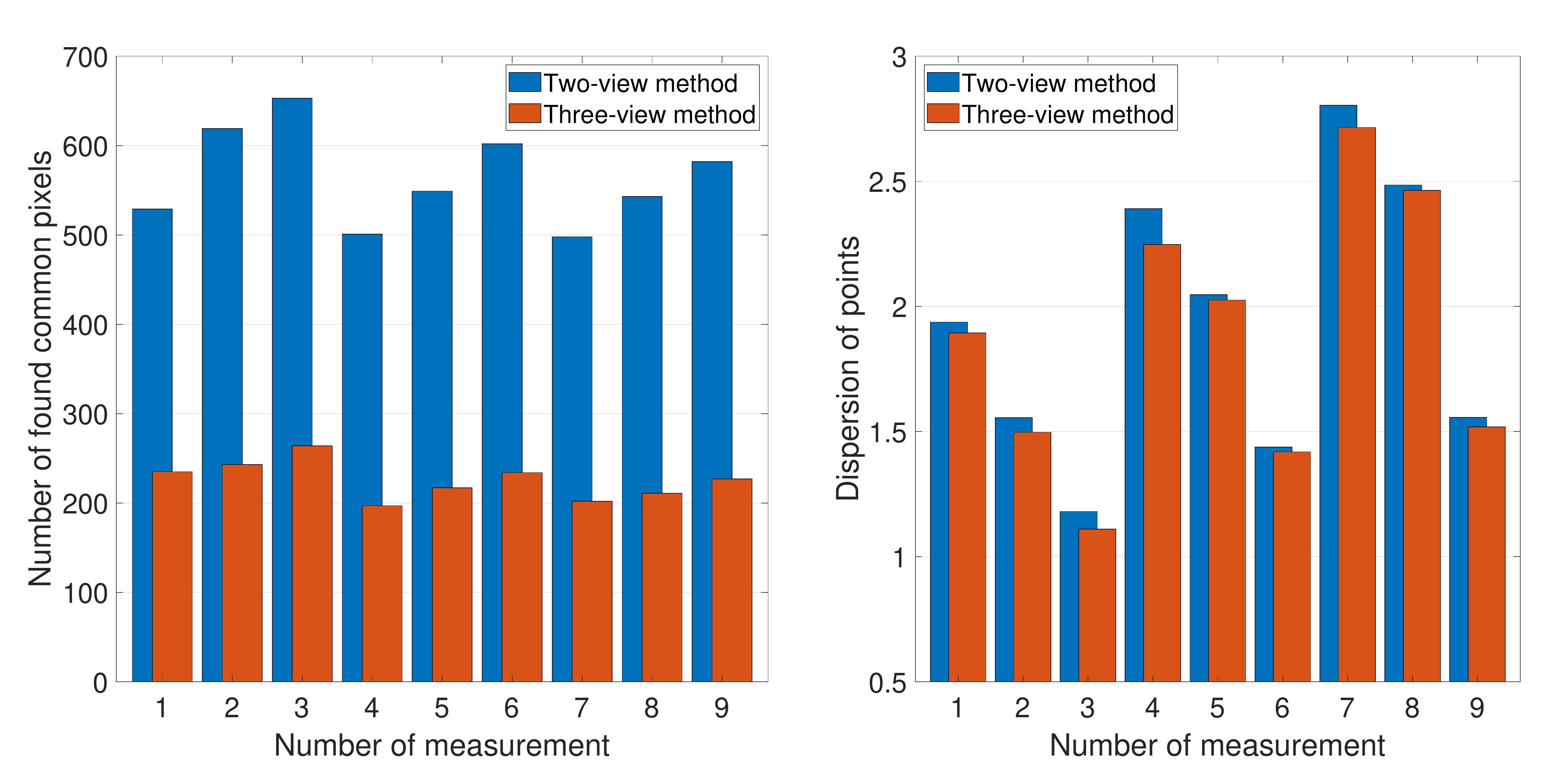}
	\caption{Different angle measurements}
	\label{fig:angles}
\end{figure}

First of all, the middle camera was fixed for all measurements at 21cm distance from the object. For every three measurement, the left camera was fixed, while the right-one was positioned in $8.74\degree, 13.18\degree$ and $16.79\degree$ degrees to the middle camera. At the first three measurements, the left camera was held in $5.33\degree$, in the next three in $9.46\degree$ and at the last three in $10.72\degree$ degrees to the middle camera (see Figure \ref{fig:ang}).
\\
The first conclusion we can observe is the resolution does not play
a key role in either two- or three-view method. For both methods,
the number of pixels matched increases in the same way as the angles
increase, but for two-view method is still significantly higher. From the current measurements, we can conclude, the dispersion decreases if one camera is positioned in a low angle, and the other in a high angle to the middle camera.  
\\
The best result is in the case when left-camera has $5.33\degree$ and the right-camera $16.79\degree$ degrees to the middle camera. In this case the number of found common pixels is $653$ for the two-view method, and $264$ for the three-view one. In this case also, the dispersion of points is minimal, $1.18$ for two-view method, and $1.11$ for three-view one. 
\\
The worst result is at $10.72\degree$ for the left camera and $8.74\degree$ for the right-one, with $498$ common pixels found using two-view method, and $202$ using three-view. The dispersions in this case are the largest, $2.8$ and $2.71$ respectively for the two-view and three-view methods.

\subsection{Evaluation of different distances measurements}

At these measurements, we fixed seven different points on a line. The points were at $1.73$ cm, $3.46$ cm, $5.33$ cm, $7.2$ cm, $9.16$ cm, and $10.93$ cm distances relatively to the first fixed point. We positioned the cameras in every combination of these points for measurements. The fourth point (with $5.33$ cm distance to the left-most point) was on the perpendicular bisector of the object.
\begin{figure}[h!]
	\centering
	\includegraphics[scale=0.25]{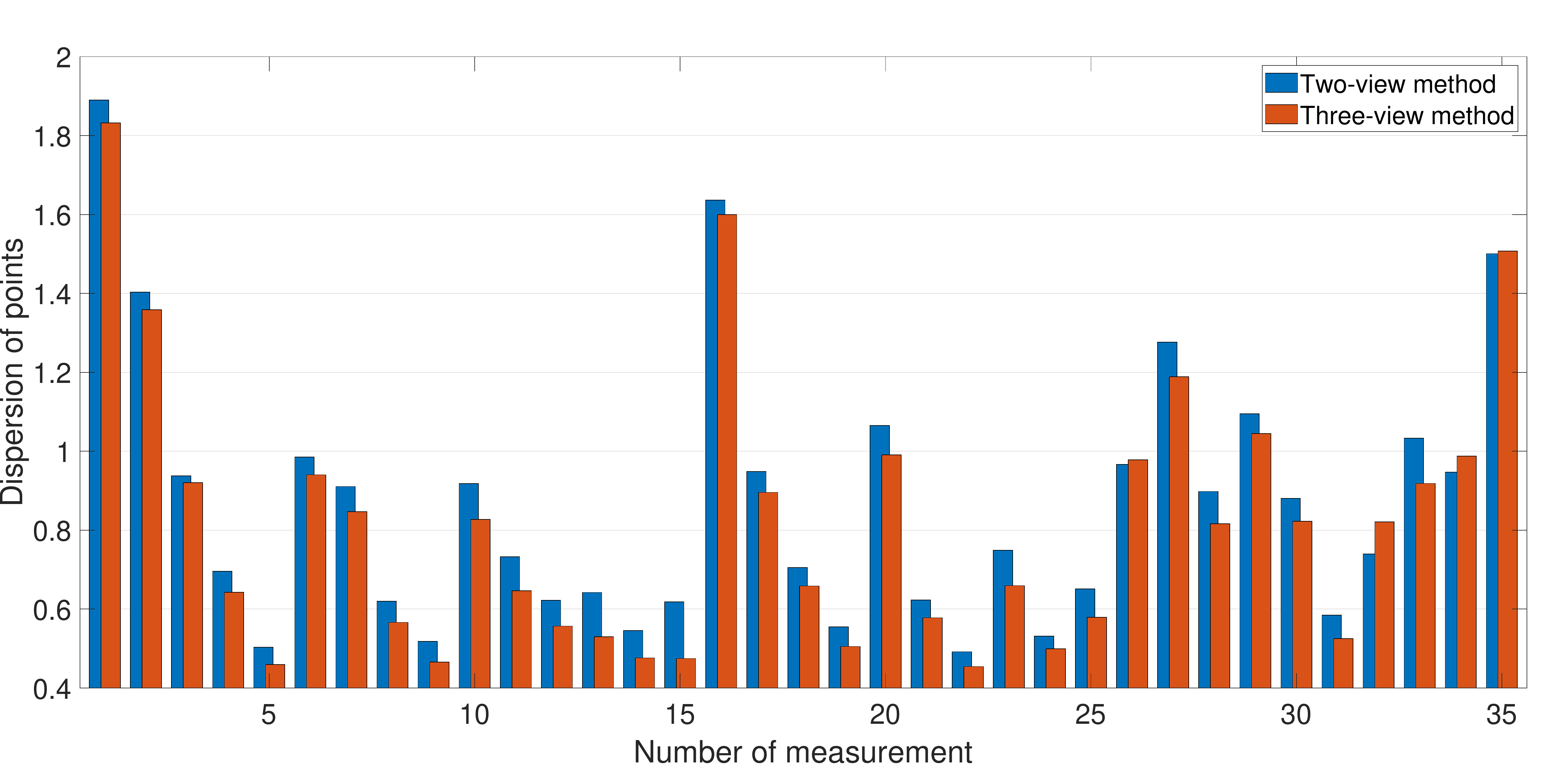}
	\caption{Different distances measurements}
	\label{fig:dist}
\end{figure}
\\
Our measurements don't show a clear correlation between the different
distances between cameras and the number reconstructed pixels, but it is clearly apparent that the dispersion decreases when two cameras are close together
and the third moves away. 
\\
The best result was when the left camera is at $1.73$ cm, the middle-one at $5.33$ cm and the right-one at $7.2$ cm distance from the left-most point. In this case the dispersion is $0.49$ for two-view method, and $0.45$ for three-view one. 
\\
The worst result was the first measurement, when the left camera is at $0$ cm, the middle-one at $1.73$ cm and the right-one at $3.46$ cm distance from the left-most point. In this case the dispersion is $1.89$ for the two-view method, and $1.83$ for the three-view one. 

\subsection{Conclusion}

From the results, firstly, we can conclude, that using the two-view method there will be more common pixels but using the three-view method the dispersion of triangulated points will be smaller. Secondly, we conclude, that the number of found common pixels increases if we increase the angle of one camera but the other camera remains at a low angle to the middle one. The dispersion of triangulated points depends on the distance between the cameras, but also from the position relative to the object. The last conclusion is that until a certain resolution, the bigger resolution results in more triangulated points. This resolution may differ from the camera used for capturing the images. In our case, the best resolution was the UltraHD resolution $(3840\times2160$ pixels$)$.  
\\
Summarizing the research output, we can formulate that using the three-view method, we get a better triangulation result for the human eye than using the two-view method. The question remains whether the three-view method is more accurate than the two-view method, or not?

\bibliographystyle{plain}
\bibliography{bibliography}

\begin{thebibliography}{10}

\bibitem{pixel2}
DeviceSpecifications.
\newblock Google pixel 2 device specifications.
\newblock \url{https://www.devicespecifications.com/en/model/cd694619}.

\bibitem{draisma2016euclidean}
Jan Draisma, Emil Horobe{\c{t}}, Giorgio Ottaviani, Bernd Sturmfels, and
  Rekha~R Thomas.
\newblock The euclidean distance degree of an algebraic variety.
\newblock {\em Foundations of computational mathematics}, 16(1):99--149, 2016.

\bibitem{openCamera}
Mark Harman.
\newblock Opencamera.
\newblock \url{https://sourceforge.net/projects/opencamera/}.

\bibitem{hartley2003multiple}
Richard Hartley and Andrew Zisserman.
\newblock {\em Multiple view geometry in computer vision}.
\newblock Cambridge university press, 2003.

\bibitem{hartley1997triangulation}
Richard~I Hartley and Peter Sturm.
\newblock Triangulation.
\newblock {\em Computer vision and image understanding}, 68(2):146--157, 1997.

\bibitem{huang2009novel}
Lili Huang and Matthew Barth.
\newblock A novel multi-planar lidar and computer vision calibration procedure
  using 2d patterns for automated navigation.
\newblock In {\em 2009 IEEE Intelligent Vehicles Symposium}, pages 117--122.
  IEEE, 2009.

\bibitem{computervision}
Zolt{\'a}n Kat{\'o} and L{\'a}szl{\'o} Cz{\'u}ni.
\newblock {\em Sz{\'a}m{\'i}t{\'o}g{\'e}pes l{\'a}t{\'a}s egyetemi tananyag}.
\newblock Typotex, 2011.

\bibitem{imgs}
Zsolt~Levente Kucsv{\'a}n.
\newblock Data set used for the research.
\newblock
  \url{https://drive.google.com/open?id=18AbYIyJLjxRKZjbxb1hyALruduCgt8qO}.

\bibitem{openMVG}
Pierre Moulon, Pascal Monasse, Renaud Marlet, and Others.
\newblock Openmvg.
\newblock \url{https://github.com/openMVG/openMVG}.

\bibitem{stewenius2005hard}
Henrik Stewenius, Frederik Schaffalitzky, and David Nister.
\newblock How hard is 3-view triangulation really?
\newblock In {\em Computer Vision, 2005. ICCV 2005. Tenth IEEE International
  Conference on}, volume~1, pages 686--693. IEEE, 2005.

\bibitem{tsai1986efficient}
Roger~Y Tsai.
\newblock An efficient and accurate camera calibration technique for 3d machine
  vision.
\newblock {\em Proc. of Comp. Vis. Patt. Recog.}, pages 364--374, 1986.

\end{thebibliography}

\bigskip
\rightline{\emph{Received:  {\tiny \raisebox{2pt}{$\bullet$\!}} Revised: }} 

\end{document}